\documentclass{article}

\usepackage{natbib}
\usepackage[final]{mlwg_2019}
\usepackage{graphicx}

\usepackage[utf8]{inputenc} % allow utf-8 input
\usepackage[T1]{fontenc}    % use 8-bit T1 fonts
\usepackage{hyperref}       % hyperlinks
\usepackage{url}            % simple URL typesetting
\usepackage{booktabs}       % professional-quality tables
\usepackage{amsfonts}       % blackboard math symbols
\usepackage{nicefrac}       % compact symbols for 1/2, etc.
\usepackage{microtype}      % microtypography
\usepackage{dsfont}
\usepackage{physics}
\usepackage[dvipsnames]{xcolor}
\usepackage{bbm}
\usepackage[toc,page]{appendix}
\usepackage[font=small,labelfont=bf]{caption}
\usepackage{multirow}
\usepackage{diagbox}
\usepackage{adjustbox}

\title{How to Incorporate Monotonicity in Deep Networks While Preserving Flexibility?}

\author{%
  Akhil Gupta\thanks{Equal contribution. Copyright 2019 by the author(s).} \\
  University of Illinois\\
  Urbana-Champaign, IL\\
  \texttt{akhilg3@illinois.edu} \\
  \And
  Naman Shukla\footnotemark[1] \\
  Deepair LLC \\
  Dallas, TX\\
  \texttt{naman@deepair.io} \\
  \And
  Lavanya Marla \\
  University of Illinois \\
  Urbana-Champaign, IL\\
  \texttt{lavanyam@illinois.edu}\\
  \And
  Arinbjörn Kolbeinsson \\
  Imperial College London \\
  London, UK\\
  \texttt{ak711@imperial.ac.uk} \\
  \And
  Kartik Yellepeddi \\
  Deepair LLC \\
  Dallas, TX\\
  \texttt{kartik@deepair.io}
}

\begin{document}

\maketitle
\vspace{-0.1cm}
% ------------- ABSTRACT -------------
\begin{abstract}
The importance of domain knowledge in enhancing model performance and making reliable predictions in the real-world is critical. This has led to an increased focus on specific model properties for interpretability. We focus on incorporating monotonic trends, and propose a novel gradient-based point-wise loss function for enforcing partial monotonicity with deep neural networks. While recent developments have relied on structural changes to the model, our approach aims at enhancing the learning process. Our model-agnostic point-wise loss function acts as a plug-in to the standard loss and penalizes non-monotonic gradients. We demonstrate that the point-wise loss produces comparable (and sometimes better) results on both AUC and monotonicity measure, as opposed to state-of-the-art deep lattice networks that guarantee monotonicity. Moreover, it is able to learn differentiated individual trends and produces smoother conditional curves which are important for personalized decisions, while preserving the flexibility of deep networks.
\end{abstract}

% ------------------------- INTRODUCTION -------------------------
\vspace{-0.1cm}
\section{Introduction}\label{introduction}
\vspace{-0.1cm}
Neural networks have recently demonstrated tremendous success in influencing decisions across mass-impact domains, such as finance \citep{7966019, AAAI1816160}, pricing \citep{CHIARAZZO2014810, shukla2019dynamic, ye2018customized} and policy-making \citep{doi:10.1080/10919392.2015.1125187}. Because decisions in these domains have significant societal implications, questions about the models' interpretability and learning behavior commonly arise. Trust in the system and associated applicability take center stage when analysts intend to validate their prior domain expertise about the data, vis-à-vis the learned statistical model. This \textit{a priori} knowledge can relate to an attribute following a certain distribution in nature. In this work, we focus on one such intuitive trend, namely, monotonicity.
%, such as unimodal, bimodal, monotonic, normal, exponential, or gamma. 

Even though real-world data encompasses high-dimensional inputs with multiple interactions \citep{HALL2014694}, it is common to possess prior domain knowledge about the monotonic trend (non-increasing / non-decreasing) between a subset of input features and the output, giving rise to partial monotonicity \citep{partial_monotone}. For instance, economic theories in house pricing \citep{house_pred} expect selling price to increase with total area. Monotonicity constraints act as a regularizer, and enhance generalization to test data, apart from facilitating human-in-the-loop adaptive learning. Such considerations demand specific attention in deep learning where prediction accuracy may be improved at the expense of interpretability \citep{chen2018learning}.

In this study, we test a novel gradient-based loss for incorporating partial monotonicity within deep neural networks, which is implemented in combination with an  empirical risk function to penalize non-monotonic trends. Our contribution is to demonstrate that monotonicity can be introduced effectively and flexibly as required by domain expertise in real-world applications. Our approach stays independent of the model structure, facilitating seamless integration with already-deployed networks. This helps us to preserve the versatility of deep networks and deliver a scalable solution. Moreover, configurable weight proportions for the two components in the custom loss help control degree of monotonicity, thereby, adding much-needed transparency across complex black-box models.

% ------------------------- RELATED WORK -------------------------
\vspace{-0.1cm}
\section{Related work}\label{related work}
\vspace{-0.1cm}
Monotonicity is enforced commonly with neural networks either by changing (i) model architecture or (ii) the learning process. Architecture alterations relate to connecting hidden nodes differently or imposing constraints on weights of certain input/hidden nodes. Most published work has focused on these structural changes - starting from positive weight constraints by \cite{wang_archer}. \cite{sillmono} introduced those constraints into a three-layer neural network for full monotonicity, which was further extended by \cite{partial_monotone} to partial monotonic functions for low-dimensional spaces. Generalizing the approach, \cite{dln} proposed deep lattice networks (DLN) using a combination of linear calibrators and lattices for learning monotonic functions. Though DLN outperforms previous methods, constructing a lattice with $D$ features ($2^D$ simplex) is computationally expensive for large spaces. Moreover, lattices, being structurally rigid, use multi-linear interpolation for unseen data - leading to step-wise and non-intuitive relationship between the input and output.

Monotonicity can be incorporated within the learning process i.e. the backpropagation step by adding constraints, either as an additional cost with standard loss function ("soft" constraint); or as a "hard" constraint, similar to Lagrangian multipliers.  \cite{mrquezneila2017imposing} compare these two types of constraints and establish that soft constraints perform better because satisfying hard constraints leads to sub-optimal outputs. \cite{pathakICCV15ccnn} show that alternative loss formulations can be equivalent to optimizing convolutional networks in a constrained manner. We contribute to this field of soft-constraint methods by introducing a point-wise loss (PWL) function for enforcing partial monotonicity within any DNN, without any change in architecture.

% ------------------------- METHODS -------------------------
\vspace{-0.1cm}
\section{Point-wise Loss (PWL)}\label{pwl}
\vspace{-0.1cm}
Consider the general setting of a supervised learning problem with a training set of $n$ examples $\{(x_i, y_i)\}$,  $i = 1, \dots , n$. The label could either be real-valued ($y_i \in \mathbb{R}$), or binary ($y_i \in \{0, 1\}$ such as classification labels). The objective of the methods presented below is to determine an estimator function $f(x)$ which is monotonic with respect to $x[M]$, where $M$ is a subset of features defined by $M \subseteq \mathcal{D}$ in $x \in \mathbb{R}^\mathcal{D}$. A function $f$ is said to be monotonically increasing in $M$ if $f(x_i) \geq f(x_j)$ for any two feature vectors $x_i, x_j \in x$, such that $x_i[M] \geq x_j[M]$ and $x_i[p] = x_j[p],$ for all $p \in \mathcal{D}\setminus M$. Without loss of generality, we use non-decreasing monotonicity for our study.

We present a point-wise loss (PWL) function that incorporates monotonic knowledge into neural networks by altering the learning process. The objective function with point-wise derivatives which embeds \textit{a priori} knowledge about monotonicity is inspired from finite element analysis as approximation \citep{strang1972approximation, wilmott1995mathematics} and classes of functions presented by \cite{dugas2009incorporating}. We formulate the following minimax objective function $\mathcal{L}_{mono}$ computed over $x_i, \; \forall \; i \in [1, n]$: %\vspace{-0.1cm}

\begin{equation}
\min \mathcal{L}_{mono} = \min \left\{ \sum_{i=1}^{n} \max \bigg(0, -{\div_M f(x_i;\theta)} \bigg) + \mathcal{L}_{NN} \right\}
\label{eq:pwl}
\end{equation}

where $\div_M$ is divergence with respect to feature set $x[M]$, i.e., $\sum_j \frac{\partial f(x_i; \theta)}{\partial x[M]_i^j} \;\; \forall\;\; j \in M$, $\theta$ are the trainable parameters, and $\mathcal{L}_{NN}$ refers to the empirical risk minimization for neural networks.

% \subsection{Deep Lattice Networks (DLN)}
% More commonly known as TensorFlow Lattice \citep{dln}, it relies on the inherent well-defined structure of an $n-dimensional$ simplex. Rigidity and relationship between any two adjacent vertices are leveraged for linear calibration of the entire feature set. The 1-D calibrators with weight constraints embed monotonicity pertaining to the desired $x[m]$ features, but the simplex interpolation having $O(\mathcal{D} \log \mathcal{D})$ complexity can take longer time in training as $D$ increases. In addition, the proposed calibration doesn't work well with sparse features as equally-spaced quantiles have been proposed. 

% Lattices are interpolated look-up tables which act as efficient nonlinear function class and can ensure monotonic behavior \citep{LR, mono_lookup}. These look-up tables are constrained and linearly calibrated to learn monotonic functions using the training data in a standard structural minimization framework. Lattice layers alternate with linear embeddings and calibrators to form the basis for a more complex structure, deep lattice network \citep{dln, EnsembleOfLattice}. This structural change guarantees monotonicity by changing the way data gets stored and retrieved (via multi-linear interpolation in $O(\mathcal{D} \log \mathcal{D})$ time). 

\vspace{-0.1cm}
\subsection{Evaluation Strategy}\label{evalStrategy}
\vspace{-0.1cm}
To evaluate the performance of DNN trained with PWL, we compare it to DLN. DLN is geared towards providing structural monotonicity guarantees irrespective of the trends in data, while PWL focuses on learning monotonicity from the data. We use two metrics: (i) area under the ROC curve (AUC) and (ii) our monotonicity metric $\mathcal{M}_k$ (defined in \eqref{eq:mono1} and \eqref{eq:mono2}) for this  analysis.

%\vspace{-0.25cm}
\begin{equation}
    \mathcal{M}_k = \frac{1}{n}\sum_{i=1}^{n} \mathbbm{1}_{(\delta_i)}, \text{where}
    \label{eq:mono1}
\end{equation}
%\vspace{-0.5cm}
\begin{equation}
    \delta_i= 
\begin{cases}
    1& \text{if } \Delta_+ f(x_j;\theta) \geq 0 \;\;\forall\;\; j \ni \{x_j[k] \in [\kappa_{min}, \kappa_{max}), x_j[p]=x_i[p] \;\;\forall\;\; p\neq k\}\\
    0              & \text{otherwise.}
\end{cases}
\label{eq:mono2}
\end{equation}

Here, the latent variable $\delta$ is an indicator that measures the degree of monotonicity $\mathcal{M}_k$ for the $k^{th}$ feature across the entire dataset. $\Delta_+$ refers to the forward difference operator for non-decreasing monotonicity \citep{wilmott1995mathematics}. The range for evaluating monotonicity of feature $k$ is set to $[\kappa_{min}, \kappa_{max})$, i.e. the lowest and highest values of the $k$th feature present in the training data.

% ------------------------- EXPERIMENTS -------------------------
\vspace{-0.1cm}
\section{Experiments}\label{ex}
\vspace{-0.1cm}
% This section sheds light on the performance of the two models, namely DLN and PWL, across the three data sources : (1) Toy dataset, (2) UCI Adult dataset and (3) Airline Ancillaries dataset. For toy dataset, we visualize the outputs on the 2-D functional space, whereas, the methods are tested on hold-out sample for other structured datasets. DLN structure is similar to the one proposed by You et al. \citep{dln} i.e. \textit{Cal-Lin-Cal-EnsLat-Cal-Lin} having two configurable parameters - $G$ (number of lattices) and $S$ (number of inputs to each lattice). For PWL, we use two fully-connected hidden layers having $H_1$ and $H_2$ units respectively. Models have been trained on cross-entropy loss, in case of classification, and squared loss for regression. PWL uses Stochastic Gradient Descent (SGD) for optimization of the objective loss function. We used the same set of monotonic features for all experiments on a dataset to maintain consistency. Datasets have been summarized in Table \ref{data}.

We compare the performance of PWL and DLN across three data sources : (1) an artificial dataset, (2) UCI - Adult dataset and (3) Airline Ancillary dataset, summarized in Appendix (Table \ref{tab:data}). Our models are trained on cross-entropy loss for classification problems, and mean squared loss for regression. PWL uses Stochastic Gradient Descent (SGD) for optimization of the objective loss function. DLN implementation is similar to the one proposed by \cite{dln}.
%We use the same set of monotonic features for all experiments on a dataset to maintain consistency.

\vspace{-0.1cm}
\subsection{Artificial dataset}\label{artificialdata}
\begin{figure}[ht]
    \centering
    \includegraphics[width=14cm]{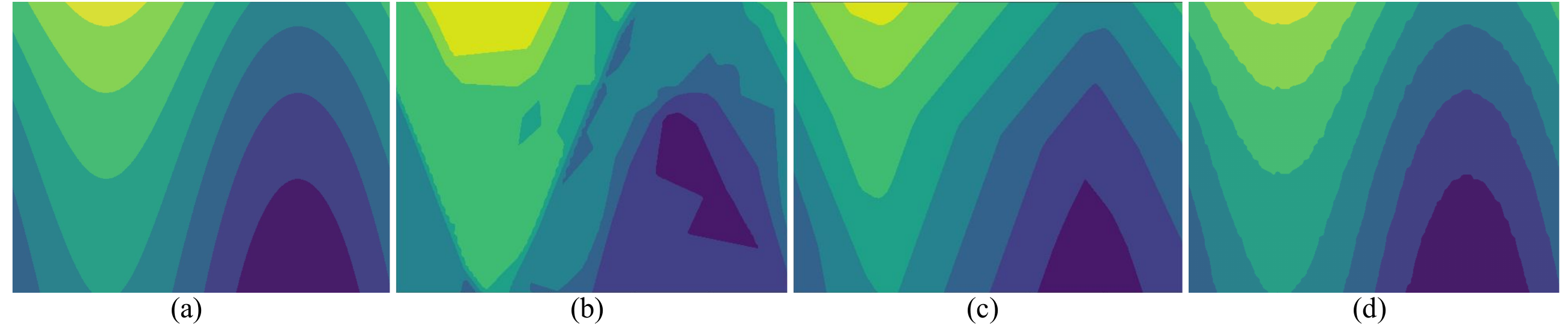}
    \caption{Function $f(x, y) =  sin(x) + e^{y}$ ; $x \in [0,1]$ and $y \in [0,1]$ are the principal axis along horizontal and vertical dimensions respectively. \textit{Contours}: (a) Target (b) DNN estimated (c) PWL estimated (d) DLN estimated}
    \label{fig:toy}
\end{figure}

\vspace{-0.1cm}
We synthesize an artificial dataset (sinusoidal in $x$ and exponential in $y$; Fig. \ref{fig:toy}a) to visualize learning on the 2-D functional space. We observe that standard DNN with squared loss and moderate batch size  detects the overall trend, but there are regions of non-monotonicity (Fig. \ref{fig:toy}b). Monotonicity is ensured along $y$ after incorporating PWL  with same DNN architecture, however, the curves appear to be marginally sharper than the target function (Fig. \ref{fig:toy}c). Contour lines for DLN  are closest to the target function (Fig. \ref{fig:toy}d). Although DLN learns better than PWL, our alternative PWL formulation is able to bring DNN closer to monotonicity without a complete overhaul in architecture, unlike DLN. 

\vspace{-0.2cm}

\begin{table}[htbp]
    \centering
  \caption{Model performance on the two datasets. ($\mathcal{M}_k$: monotonicity metric \eqref{eq:mono1};  $\mathcal{T}$: run-time in $10^3$ seconds)}
  \begin{tabular}{c | c c c | c c c }
    \hline
    \multirow{2}{*}{Models} &
      \multicolumn{3}{c}{UCI - Adult} &
      
      \multicolumn{3}{c}{Airline Ancillary} \\

    & AUC & $\mathcal{M}_k$ & $\mathcal{T}$ & AUC & $\mathcal{M}_k$ & $\mathcal{T}$\\
    \hline
    DLN & \textbf{0.917} & \textbf{1.000} & 5.586 & 0.708 & \textbf{1.000} & 7.770\\
    
    PWL & 0.908 & 0.856 & \textbf{0.338} & \textbf{0.722} & 0.985 & \textbf{1.375}\\
    \hline
  \end{tabular}
  \label{tab:results}
\end{table}

\vspace{-0.1cm}
\subsection{UCI - Adult dataset}\label{ucidata}
\begin{figure}[ht]
    \centering
    \includegraphics[width=10cm]{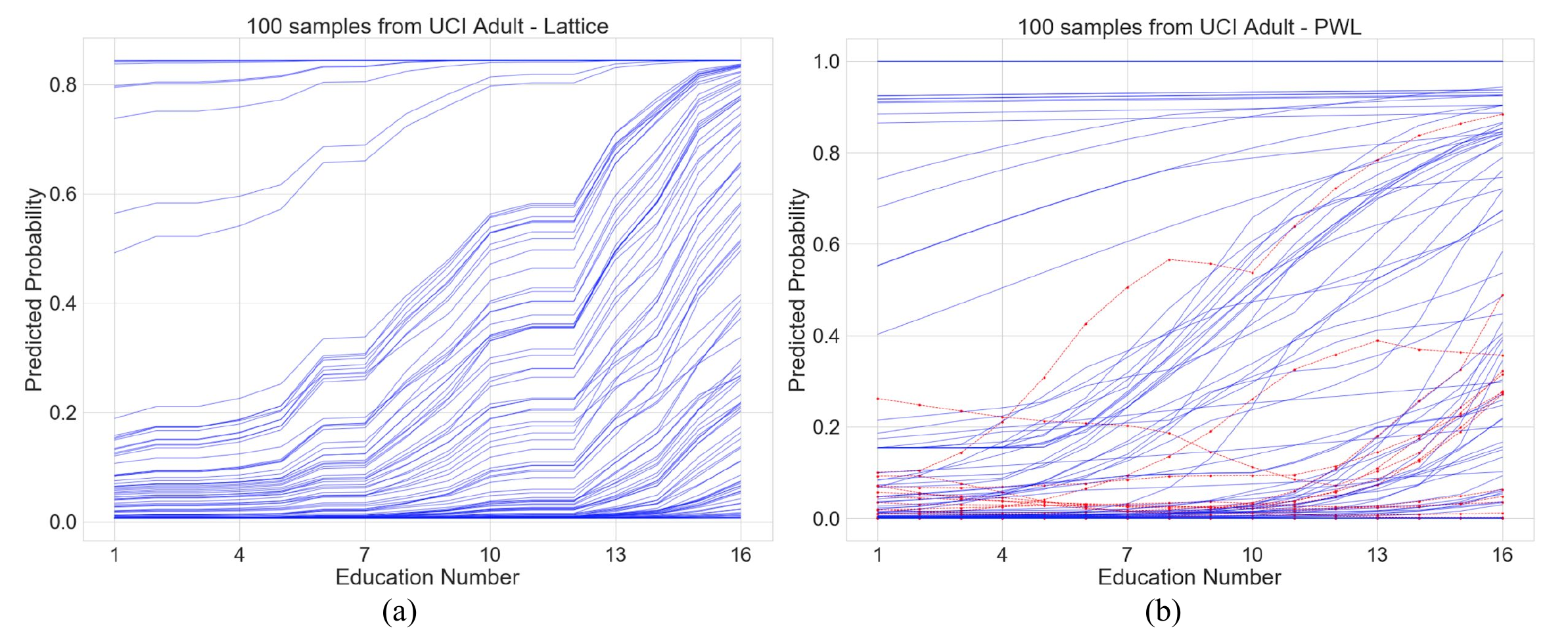}
    \caption{UCI - Adult dataset: Conditioned trends for Education Level}
    \label{fig:adult}
\end{figure}

\vspace{-0.1cm}
Similar to \cite{dln}, census data from the UCI repository \citep{adult_df} is used to predict whether a person's income exceeds \$50,000 or not. Monotonicity is enforced with respect to education level, hours per week, and capital gain. Considering education level for illustration, we note that DLN marginally edges out PWL on AUC and monotonicity measure $\mathcal{M}_k$ (See Table \ref{tab:results}). However, analysis of conditioned trends on education level (Fig. \ref{fig:adult}a) suggests that the DLN is learning similar (consistent step-wise slope) patterns for each person. PWL results demonstrate varying trajectories for each person, i.e. it differentiates between individuals after considering non-linear interactions. We found the actual correlation between education level and income to be $0.33$, indicating that the data may not be completely positive monotonic. Hence, the DLN does guarantee monotonicity but at the cost of sometimes ignoring real signals from data, which are detected by PWL (red lines in Fig. \ref{fig:adult}b).

%\vspace{-0.25cm}
%\vspace{-0.1cm}
\begin{figure}[ht]
    \centering
    \includegraphics[width=10cm]{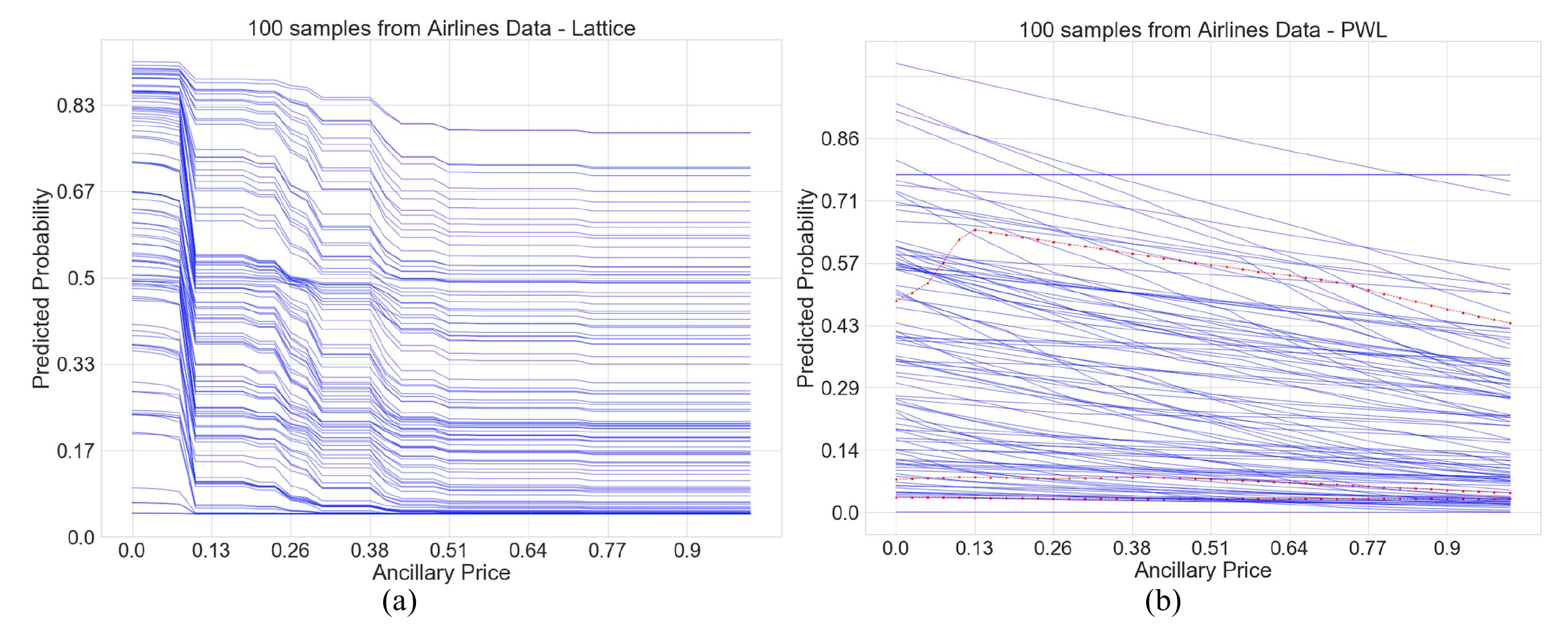}
    \caption{Airline Ancillary dataset: Conditioned trends for Ancillary Price}
    \label{fig:price}
\end{figure}
%\vspace{-0.15cm}

\vspace{-0.1cm}
\subsection{Airline Ancillary dataset}\label{airlinedata}
\vspace{-0.1cm}
Ancillary pricing \citep{shukla2019dynamic, shukla2019adaptive} is a sub-field within airline pricing where ancillaries are priced in association with partner airlines using an A/B testing framework. Domain knowledge suggests that the ancillary purchase probability should follow a non-increasing monotonic trend with respect to price. We use this dataset to evaluate the DLN and PWL model performances in a real-world setting. Upon experimenting with different iterations of the DLN i.e. increasing lattice size and number of lattices, we find that DLN's training time increases significantly without noticeable improvement in AUC (Appendix: Table \ref{tab:res-perf}). On the other hand, PWL model is more tractable and solves faster (Table \ref{tab:results}). In addition, we find that PWL outperforms DLN in terms of AUC and has similar $\mathcal{M}_k$ score (Table \ref{tab:results}). DLN makes predictions in a step-wise fashion with linearly interpolated segments (Fig. \ref{fig:price}a), thereby missing out on customer dynamics for unobserved regions. However, PWL produces curves with smoother derivatives - suggesting the possibility of greater personalization for customers, and establishing continuity of trend on either side of the price horizon (as expected).

% ------------------------- CONCLUSION -------------------------
\vspace{-0.1cm}
\section{Conclusion}
\vspace{-0.1cm}
In this work, we explored one way of incorporating \textit{a priori} knowledge - monotonicity, to leverage domain expertise into data-driven approaches. We tested a point-wise loss (PWL) function that penalizes non-monotonicity, against deep lattice networks (DLN) which enforce monotonicity via calibrated look-up tables. We discussed how DLN guarantees monotonicity at the cost of learning from data, whereas PWL strives to achieve a compromise between minimizing empirical risk and enforcing monotonicity. DLN is unable to differentiate between individual data points, a model characteristic desired in the real-world to drive personalized decisions. In contrast, varying trajectories learned by PWL provide more power for customized decision-making. In addition, PWL can be used as a plug-in for enforcing soft monotonicity, while retaining the flexibility and power of deep networks. However, PWL demands a wise choice of the relative importance of different terms in the loss function - an open problem in the deep learning community. In future work, we aim to facilitate transfer learning between different PWL model iterations to regularize the loss weight proportions.
%We shall also explore networks with hard constraints on model outputs to benchmark the PWL's performance.

\subsubsection*{Acknowledgments}
We sincerely and gratefully acknowledge our airline partners for their continuing support. The academic partners are also thankful to
deepair (\href{https://www.deepair.io/}{www.deepair.io}) for funding this research.

% Use unnumbered third level headings for the acknowledgments. All acknowledgments
% go at the end of the paper. Do not include acknowledgments in the anonymized
% submission, only in the final paper.

\bibliographystyle{unsrtnat}

%\bibliography{refs}

\newpage
\begin{appendices}

\section{Additional Information}
Table \ref{tab:data} provides additional details on the experimental setup mentioned in Section \ref{ex}. Function approximation on artificial dataset presented in Section \ref{artificialdata} has been formulated as a regression task. Hence, we used mean squared error as loss function for simplicity. Cross-entropy loss was used as the primary loss function for UCI - Adult and Airline Ancillary datasets. Here, loss function refers to $\mathcal{L}_{NN}$ defined in Equation \ref{eq:pwl}.

\begin{table}[h]
  \caption{Data Summary}
  \label{tab:data}
  \centering
  \begin{tabular}{ccccc}
    \toprule
    Dataset & Type & \# Training & \# Test & \# Features (Monotonic) \\
    \midrule
    Artificial & Regression & 10,000 & - & 2 (1)\\
    UCI - Adult & Classification & 26,048 & 16,281 & 90 (3)\\
    Airline Ancillary & Classification & 106,003 & 26,513 & 32 (1) \\
    \bottomrule
  \end{tabular}
\end{table}

In Table \ref{tab:res-perf}, parameter $\theta_1$ refers to number of monotonic lattices and $\theta_2$ refers to the lattice size/rank for DLN. For DNN based models trained with PWL, these parameters are the hidden units in the 2-layer neural network. Performance has been evaluated on a machine with NVIDIA Tesla K80 GPU. 

\begin{table}[h]
  \caption{Comparison of various experiments on PWL and DLN.}
  \label{tab:res-perf}
  \centering
  \begin{tabular}{cccccc}
    \toprule
    Dataset & Model & Parameters [$\theta_1, \theta_2$]& AUC & $\mathcal{M}_k$ & $\mathcal{T}_{10^3}$\\
    \midrule
    UCI - Adult & DLN & [70, 5] & \textbf{0.917} & 1.000 & 5.586\\
    UCI - Adult & PWL & [32, 11] & 0.908 & 0.856 & 0.338\\
    Airline Ancillary & DLN & [20, 3] & 0.709 & 1.000 & 0.277 \\
    Airline Ancillary & DLN & [30, 3] & 0.706 & 1.000 & 0.373 \\
    Airline Ancillary & DLN & [70, 3] & 0.708 & 1.000 & 0.738 \\
    Airline Ancillary & DLN & [70, 4] & 0.709 & 1.000 & 1.451 \\
    Airline Ancillary & DLN & [70, 5] & 0.708 & 1.000 & 7.770\\
    Airline Ancillary & PWL & [32, 11] & \textbf{0.722} & 0.985 & 1.375\\
    \bottomrule
  \end{tabular}
\end{table}

\section{Additional Observations}

Various iterations of DLN on airline ancillary dataset (Section \ref{airlinedata}) show that an increase in lattice size/rank ($\theta_2$) for enhanced model learning leads to an exponential increase in the run-time, without any improvement in AUC (Table \ref{tab:res-perf}). On the other hand, PWL converges faster and outperforms DLN on the AUC. Even though $\mathcal{M}_k$ for PWL is marginally lower than 1 (DLN), PWL is picking the real signal from the data when sample behavior is not monotonic in practice.

\begin{figure}[ht]
    \centering
    \includegraphics[width=12cm]{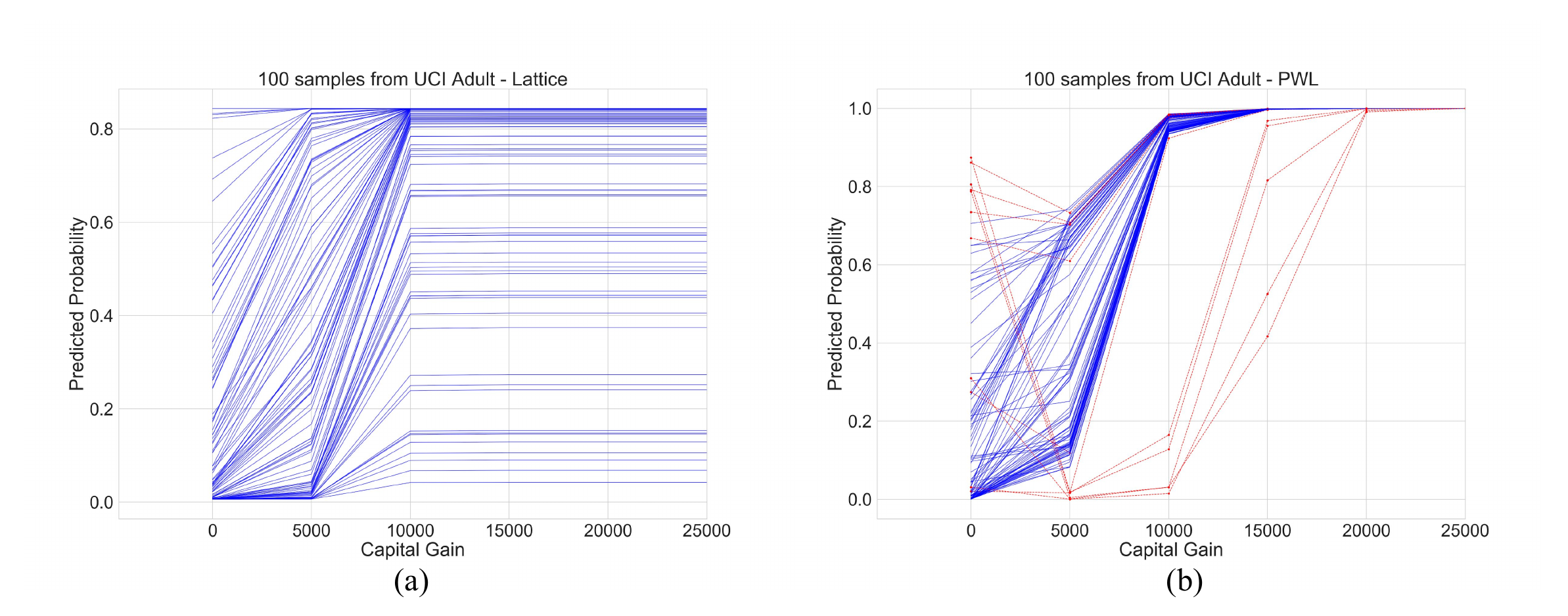}
    \caption{UCI - Adult dataset: Conditioned trends for Capital Gain}
    \label{fig:cg_adult}
\end{figure}

Model results for PWL and DLN on the UCI - Adult dataset (capital gain feature) are presented in Figure \ref{fig:cg_adult}. For this feature, DLN shows more variance in learning curves than PWL which generates similar trajectory for a group of data points. Reason can be attributed to the interdependence of different loss terms in the PWL function, necessitating the importance of balancing trade-off. For instance, $(\mathcal{L}_{mono} - \mathcal{L}_{NN})$ i.e. the monotonicity loss could be decreasing with respect to one of the monotonic feature but not getting equally minimized for the other monotonic feature. One way of countering this limitation is transfer of model weights between various PWL iterations for optimization of the custom loss coefficients (weights of $\mathcal{L}_{mono}$ as defined in Equation \ref{eq:pwl} - kept same for this work).

\begin{figure}[ht]
    \centering
    \includegraphics[width=12cm]{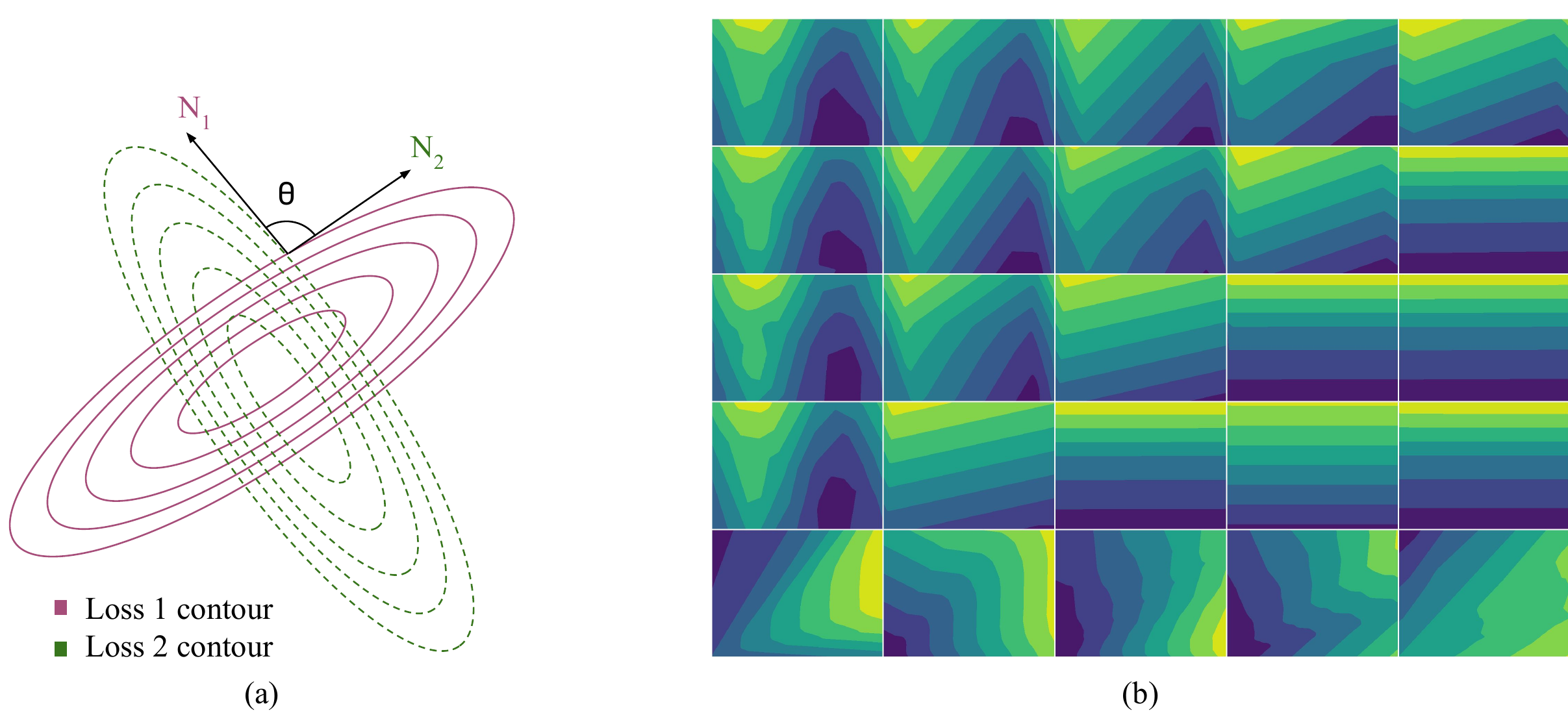}
    \caption{Artificial dataset analysis. (a) Illustration of contour plots for two competing loss functions: loss 1 and loss 2 with respective normal vectors as $N_1$ and $N_2$. (b) PWL estimated contour plot for different hyperparameters: horizontally increasing frequency for $(\mathcal{L}_{mono} - \mathcal{L}_{NN})$ from left to right, vertically increasing frequency for $\mathcal{L}_{NN}$ from bottom to top.}
    \label{fig:ill}
\end{figure}

The weighted loss method or the loss term switching method can be used to train PWL-based model. Figure \ref{fig:ill}b shows the estimated contour plots for different frequencies (hyperparameter) when switching between two objective functions. As we move from left to right, $(\mathcal{L}_{mono} - \mathcal{L}_{NN})$ has an overpowering effect and flat lines are observed. Hence, hyperparameter is crucial for maintaining equilibrium between losses. Figure \ref{fig:ill}a shows a graphical illustration for two different loss contour planes which are mutually inclined at an angle $\theta$ with respect to each other. If the contours are not perpendicular $(\theta \ne 90^{\circ})$, the SGD steps for each of the loss values could be sub-optimal. This may result in instability while training PWL and can be considered as one of the limitations. We plan to study the loss trade-off and sub-optimality analysis in future. 

\end{appendices}
\end{document}